\documentclass{article}
\usepackage{spconf,amsmath,graphicx}
\usepackage{booktabs}
\usepackage{multirow}
\usepackage{pifont}
\usepackage{siunitx}
\usepackage{acronym}

\newcommand{\xmark}{\ding{55}}%
\usepackage{enumitem}
\setlist{nosep, leftmargin=14pt}

\usepackage{mwe} % to get dummy images
\title{TriFormer: A Multi-modal Transformer Framework For Mild Cognitive Impairment Conversion Prediction}
\name{Linfeng Liu$^{1, \dagger}$\sthanks{Corresponding author: L. Liu: linfeng.liu@uq.edu.au}, Junyan Lyu$^{1,3}$\sthanks{L.Liu and J.Lyu contributed equally to this paper}, Siyu Liu $^{2}$, Xiaoying Tang$^{3}$, Shekhar S. Chandra$^{2}$, Fatima A. Nasrallah$^{1}$}

\address{$^{1}$ Queensland Brain Institute, University of Queensland, Brisbane, Australia \\$^{2}$ School of Information Technology and Electrical Engineering, \\University of Queensland, Brisbane, Australia \\$^{3}$Department of Electronic and Electrical Engineering, \\Southern University of Science and Technology, Shenzhen, China}
\begin{document}
% \ninept
%
\maketitle
%
% \begin{abstract}
% The prediction of mild cognitive impairment (MCI) conversion to Alzheimer's disease is important for early treatment to prevent or delay the progression of AD. To efficiently predict the conversion of MCI patients, we propose a novel transformer-based framework, named TriFormer which extracts multi-view image features and exploits the correlation between various clinical data with two transformer-based modules and then fuses multi-modal features with a modality fusion transformer. We quantitatively evaluate our model on the ADNI1 and ADNI2 datasets. The proposed TriFormer outperforms previous state-of-the-art models using either single or multi-modal data by a considerable margin. 
% % The code for TriFomer is available at: https://github.com/LinfengLiu98/TriFormer.
% \end{abstract}

\begin{abstract}
The prediction of mild cognitive impairment (MCI) conversion to Alzheimer's disease (AD) is important for early treatment to prevent or slow the progression of AD. To accurately predict the MCI conversion to stable MCI or progressive MCI, we propose TriFormer, a novel transformer-based framework with three specialized transformers to incorporate multi-modal data. TriFormer uses I) an image transformer to extract multi-view image features from medical scans, II) a clinical transformer to embed and correlate multi-modal clinical data, and III) a modality fusion transformer that produces an accurate prediction based on fusing the outputs from the image and clinical transformers. Triformer is evaluated on the Alzheimer’s Disease Neuroimaging Initiative (ADNI) 1 and ADNI2 datasets and outperforms previous state-of-the-art single and multi-modal methods.

\end{abstract}
\acrodef{ADNI}{Alzheimer’s Disease Neuroimaging Initiative}
\acrodef{CDRSB}{Clinical Dementia Rating Sum of Boxes}
\acrodef{ADAS}{Alzheimer's disease assessment scale}
\acrodef{MMSE}{Mini-Mental State Exam}
\acrodef{RAVLT}{Rey Auditory Verbal Learning Test}
\acrodef{AD}{Alzheimer's disease}
\acrodef{CN}{Congnitively normal}
\acrodef{sMCI}{stable MCI}
\acrodef{pMCI}{progressive MCI}
\acrodef{MCI}{Mild cognitive impairment}
\acrodef{MRI}{Magnetic resonance imaging}
\acrodef{PET}{Positron emission tomography}
\acrodef{ViT}{Vision Transformer}
\acrodef{CNN}{Convolutional neural network}
\acrodef{MLP}{multi-layer perceptual}
\section{INTRODUCTION}
\label{sec:intro}
% Points to talk about: 1. what is MCI and why its important? 2. A little bit of introduction to AD. 3. Importance of MRI using to detect AD but not easy to detect MCI. Therefore, we need to use multi-model data. 4. Our model address....
\ac{MCI} patients exhibit a memory impairment earlier than the expected age. It is a transitional stage from \ac{CN} to \ac{AD} where around 10\% to 15\% \ac{MCI} patients end up progressing to \ac{AD} every year \cite{dunne2021mild}. Patients having \ac{MCI} can either progress to \ac{AD} within several years defined as \ac{pMCI} or stay at the same \ac{MCI} stage defined as \ac{sMCI}. Previous studies have shown that early nonpharmacological therapy and treatment can delay the progression from \ac{MCI} to \ac{AD}. However, the prerequisite to early intervention is accurately predicting the likelihood of \ac{MCI} conversion to \ac{AD} for \ac{MCI} patients at the early stage. Utilizing multi-modal clinical data such as cognitive test results, genetic information, and imaging data including T1w or T2w \ac{MRI} and \ac{PET} could help more accurately predict \ac{MCI} conversion \cite{guan2021mri}. 
% Recently, studies have shown that the usage of these multi-modal data can significantly improve the performance of \ac{MCI} prediction and \ac{AD} classification \cite{guan2021mri, pan2020spatially, qiu2020development, lian2018hierarchical}.

\acp{CNN} have been widely applied to AD classification and prediction from imaging data. Valliani et al. \cite{valliani2017deep} fine-tuned a pretrained ResNet-50 to classify \ac{AD} and \ac{CN} based on 2D axial slices. Wen et al. \cite{wen2020convolutional} leveraged 3D spatial information by using a 3D CNN and outperformed previous 2D-based methods in AD classification and \ac{MCI} conversion prediction. However, both 2D and 3D \acp{CNN} have a strong inductive bias towards local receptive fields, which could limit the performance on high dimensional data \cite{Jang_2022_CVPR}. Recently, transformers have been shown to be effective in capturing global long-range dependency within imaging \cite{dosovitskiy2020image} and sequential data \cite{devlin2018bert}. They also have no indictive bias compared with \acp{CNN}.  Despite the performance advantages, few studies have attempted transformers for predicting the progression of \ac{MCI} to \ac{pMCI}/\ac{sMCI} or classifying \ac{AD}. This is mainly due to their prohibitive computational costs on 3D medical imaging data. A novel tractable transformer-based network is needed to enhance the performance of \ac{AD} classification and \ac{MCI} conversion predictions. Furthermore, several studies have shown that using multi-modal data can significantly improve the accuracy of \ac{MCI} conversion prediction and \ac{AD} classification. Qiu et al. \cite{qiu2020development} used multi-modal data including age, gender, \ac{MMSE} and \ac{MRI} to diagnose \ac{AD} using patch-based 3D \ac{CNN}. Pan et al. \cite{pan2020spatially} proposed a spatially-constrained Fisher representation network for \ac{AD} diagnosis using \ac{MRI} and \ac{PET} imaging data. Guan et al. \cite{guan2021mri} proposed a knowledge distilling network using multi-modal data for \ac{MCI} conversion prediction. These works typically used a simple \ac{MLP} network to weight different clinical data. To the best of our knowledge, there is no existing multi-modal work adopting a transformer to explicitly leverage the correlation between clinical features for more accurate \ac{MCI} predictions.

In this paper, we propose a novel transformer-based network, TriFormer, which is made up of three transformers to exploit multi-model data including T1 weighted \ac{MRI} and clinical data to predict \ac{MCI} conversion. Our contributions are summarized as follows: 1. We propose a 2.5D \ac{ViT} to efficiently extract multi-view image features. 2. With our proposed clinical transformer, TriFormer becomes the first transformer to embed and correlate between different clinical features for \ac{MCI} conversion prediction. 3. We further propose a modality fusion transformer to aggregate the multi-modal features from the clinical and image transformers. The three transformers are put together to construct Triformer, which improves the performance of \ac{MCI} conversion prediction yielding state-of-the-art results.

\label{sec:literature_review}

\section{METHODS}
As shown in Figure 1, the proposed TriFormer contains three transformer modules: (1) an image feature extractor containing a 2.5D \ac{ViT} to extract global image features from coronal, sagittal, and axial views, (2) a second transformer to distill information from clinical data, (3) a third modality fusion transformer which aggregates multi-modal representations and provide final predictions.

\begin{figure}
    \centering
    \includegraphics[width=1.0\columnwidth]{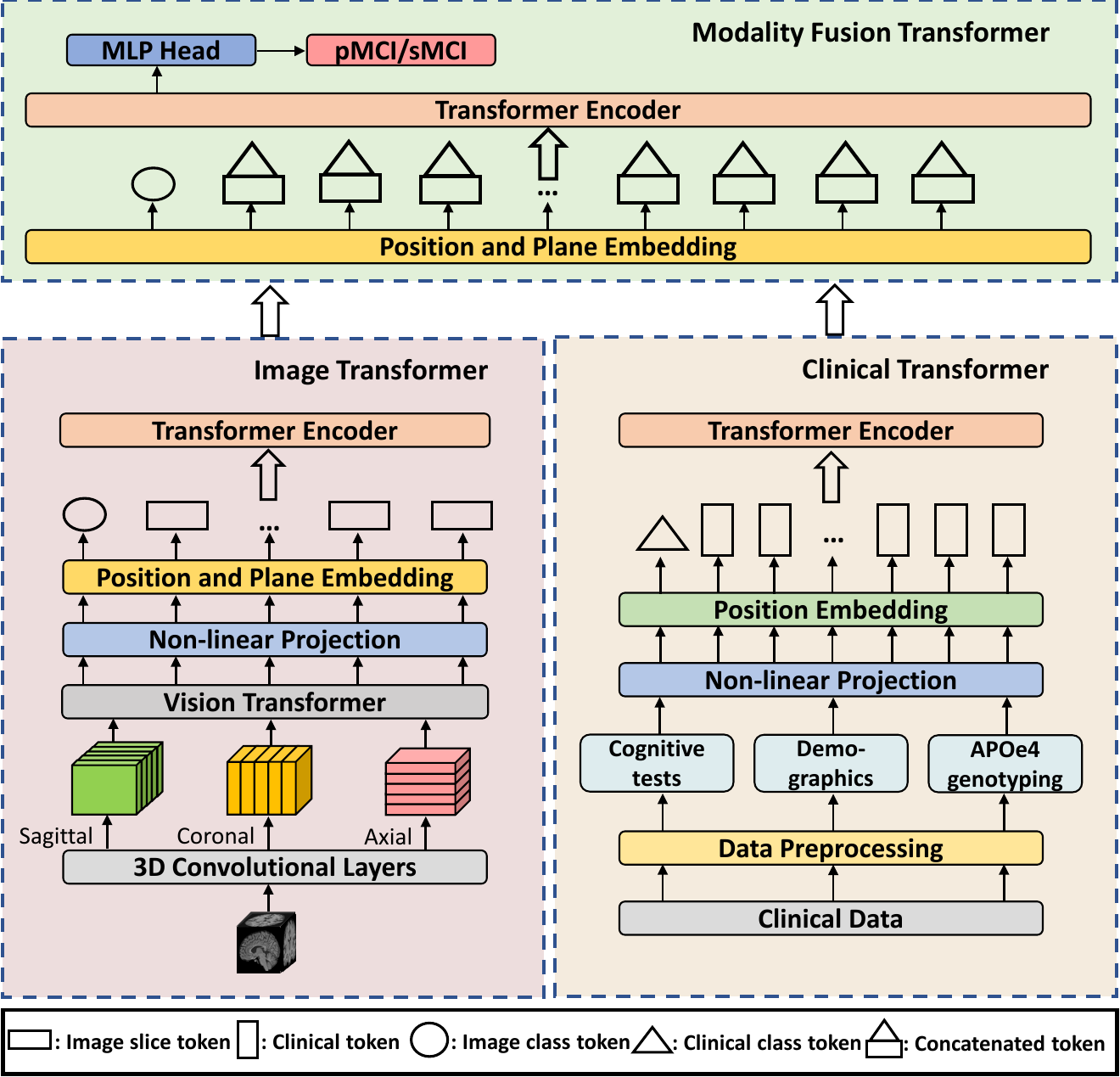}
    \caption{TriFormer architecture overview. The image transformer on the left extracts multi-view image features from MRI using ViT. The clinical transformer on the right studies the correlation between different clinical data. Image slice tokens are concatenated with the clinical class token and act as the input of the modality fusion transformer, which combines the extracted multi-modal features to perform more accurate MCI conversion predictions.}
    \label{fig:model_archi}
\end{figure}

\subsection{Image Feature Extractor}
Given a \ac{MRI} image with dimension $H \times W \times D$, the feature embedding layer which contains two 3D convolutional layers project input into a C-dimensional latent space. Then the extracted volumes are sliced into 2D along the three spatial dimensions For instance, we obtain $H$ slices of features with the dimension of $W \times D \times C$ from the axial view. As a result, concatenate them to $(H + W + D)$ slices of image features. Compared to directly feeding the raw input image slices into the module, feature embedding helps preserve more spatial information from neighbouring slices. We then adopt \ac{ViT} \cite{dosovitskiy2020image} as our feature extraction module to explore the global semantic information from the embedded features and encode them into compact feature vectors. Specifically, image features are fed into a shared-parameter \ac{ViT} slice by slice predicting $(H + W + D) \times C^{'}$ feature vectors.
Following the image feature extraction stage, we apply another transformer encoder to explore the correlation between feature vectors from different planes and slices. In this stage, we introduce a novel positional embedding layer to inject the plane information. Instead of using sinusoidal positional encoding, we provide a unique identifier for each plane to make the transformer location-aware. A plane separation token is further used for the model better to differentiate the different views. The final input to the transformer encoder is formulated as:
\begin{equation}
    \begin{aligned}
      I_{fe} = [I_{cls}, F^{cor}_1, ... F^{cor}_H, I_{sep}, F^{sag}_1,...,\\ F^{sag}_W, I_{sep}, F^{ax}_1, ..., F^{ax}_D, I_{sep}],     
    \end{aligned}
\end{equation}
where $I_{fe}$ is the image feature, $I_{cls}$ is the image class token, $I_{sep}$ is the plane separation token, $F^{cor}_H$, $F^{sag}_W$, $F^{ax}_D$ are the image slice features from all three views.
\begin{align}
I_{pos} = [I^{pos}_1,  I^{pos}_2, ..., I^{pos}_{H + W + D + 4}],
\end{align}
where $I_{pos}$ is the position embedding of each token.
\begin{align}
I_{pl} = [0, I^{cor}_1, ...,I^{cor}_H, 0, I^{sag}_1, ..., I^{sag}_W, 0, I^{ax}_1, ..., I^{ax}_D, 0],
\end{align}
where $I_{pl}$ is the plane feature with the view information of where each image slice token is from.
\begin{align}
I = I_{fe} + I_{pos} + I_{pl}.
\end{align}
All the features and embeddings are added together before feeding into the image transformer encoder.

\subsection{Clinical Feature Extractor}
Each clinical modality is first normalized between 0 and 1 and passed through a non-linear projection layer containing an MLP layer followed by a ReLU activation function. Then all clinical modalities are stacked together and passed through a learnable positional embedding layer with a clinical class token prepended before feeding into the transformer encoder. Instead of separately inputting each clinical feature to the transformer encoder of the modality fusion transformer, the clinical class token is used to inject into all the clinical features at once using a single vector.
%Normalize
% can be denoted as $M = [E_{cls}, M_1, ..., M_N] + P_{pos}$. The clinical feature $M$ is the input of the transformer encoder $T_c$ which has the same components as $T_i$. The output from $T_c$ can be denoted as $T_{c_{out}}^ {N \times d}$.

\subsection{Modality Fusion Transformer}
The clinical class token is first duplicated $H + W + D + 4$ times and each concatenated with an image slice token from the image feature extractor. These aggregated features are then fed into a modality fusion transformer which has the same positional and plane embedding layers as the image transformer. Lastly, the class token is passed through \ac{MLP} layers to perform a binary classification between \ac{pMCI} and \ac{sMCI}.
% The fusion transformer takes the extracted image and clinical features and aggregated them together to output \ac{MCI} conversion prediction. The class token of $T_{c_{out}}$ is first copied $H + W + D + 4$ time and concatenated with $T_{i_{out}}^{(3N+4) \times 2d}$ denoted as $T_{f_in}$. The positional encoding for $T_f$ also has plane encoding and a normal positional encoding layer like $T_i$.  

\section{RESULTS AND DISCUSSION}
\subsection{Dataset}
We conducted our experiments on two public datasets, the \ac{ADNI}1 and \ac{ADNI}2. \ac{ADNI}3 was not included due to the lack of \ac{MCI} patients with sufficient follow-up timepoints. All the T1-weighted images were first reoriented to the standard space and cropped automatically using FSL. Non-linear registration to the MNI152 template, skull stripping and bias field correction were applied for data normalization. Finally, all images were resampled to 128x128x128 with a voxel size of $1 \times 1 \times 1 mm^3$ using ANTs. Before feeding into TriFormer, all image intensity values were normalized between 0 and 1. Only baseline images were included in this study and the data split was based on the patient level to avoid data leakage. We included 12 types of clinical data: age, gender, education years, APOe4 genotyping, \ac{CDRSB}, \ac{ADAS} (ADAS11, ADAS13), \ac{MMSE} score and \ac{RAVLT} (immediate, learning, forgetting, percent forgetting). The subjects were separated into three groups: \ac{AD}, \ac{CN} and \ac{MCI}. The \ac{MCI} group was subdivided into two groups: \ac{sMCI} and \ac{pMCI}. Patients were diagnosed as \ac{MCI} at all available timepoints with at least 24 months records were defined as \ac{sMCI} while patients were diagnosed as \ac{MCI} at baseline but converted to \ac{AD} within 36 months without changing back to \ac{CN} or \ac{MCI} at all available timepoints were defined as \ac{pMCI}. Details of the demographics of patients are shown in Table 1.
\begin{table}\label{demographics}
\caption{Demographic information of study data}
\vspace*{1mm}
\resizebox{\columnwidth}{!}{
\begin{tabular}{lcccl}\toprule
Dataset & Label & No. of Subjects & Age Range & Male/Female\\\midrule
\multirow{2}{*}{\ac{ADNI}1} & pMCI & 137 & 55.2-88.3 & 81/56\\
{} & sMCI & 100 & 57.8-87.9&60/40\\
{} & AD & 169&55.1-90.1& 87/82\\
{}& CN & 206&59.9-89.6& 103/103\\
\multirow{2}{*}{\ac{ADNI}2} & pMCI & 57 & 55.0-84.6 & 30/27\\
{} & sMCI & 117 & 55.9-91.3&65/52\\
{} & AD & 102&55.9-88.3& 58/44\\
{}& CN & 137&56.2-85.6& 67/70

\\\bottomrule
\end{tabular}
}
\end{table}

\subsection{Implementation}

TriFormer is implemented in Pytorch and trained on an NVIDIA Tesla V100 GPU. Data augmentation methods including Random corona-view flipping and gaussian noising are applied during the training. We also utilize \ac{AD} and \ac{CN} patients to augment training data considering \ac{MCI} group is small as previous studies have shown learning from \ac{AD} and \ac{CN} can improve the \ac{MCI} prediction performance \cite{guan2021mri}. TriFomer is trained for 50 epochs using the Adam optimizer with a batch size of 2 and the cross-entropy loss. All three transformers contain 6 transformer layers with 8 heads. Image dimensions $H, W, D$ are 128 each and $C, C'$ are 32 and 512, respectively. All the embedding layers' dimensions are 256. We first train TriFormer on the \ac{ADNI}1 dataset and evaluate it on the \ac{ADNI}2 dataset, then we swap the two datasets to train on \ac{ADNI}2 and evaluate on the \ac{ADNI}1 dataset. An 80\%/20\% train-validation is applied to each experiment. The weights with the highest AUC value on the validation set is selected as the final weights. We report the average results across 5 repeats to reduce variance in the experimental results.

\subsection{Comparisons with state-of-the-art methods}
Table 2 compares TriFormer's area under the curve (AUC) and accuracy (ACC) metrics with other state-of-the-art (SOTA) methods. The results indicate TriFormer can effectively utilize multi-modal data to outperform single-modality works such as \cite{Jang_2022_CVPR}. Distinct improvements can also be observed when comparing TriFormer to the previous \ac{CNN}-based multi-modal works \cite{ guan2021mri, qiu2020development, pan2020spatially, lin2018convolutional}. Finally, the TriFormer also outperforms another transformer \cite{zheng2022transformer} that fuses cortical features from \ac{MRI} in predicting \ac{MCI} conversion. The proposed TriFormer network outperforms previous SOTA methods on both ADNI1 and ADNI2.
% To have a comprehensive comparison with other works, we compared works using only \ac{MRI} as a single modality, multi-modal imaging data (MRI + PET), and multi-modal working with \ac{MRI} and clinical data. Most of the listed works used 36 months to define sMCI and pMCI and we applied the same criteria for a fair comparison. We also collected the second most MCI patients in our study excluding Pan et al's work as they used \ac{MRI} and \ac{PET} imaging. The proposed method has a significant improvement compared to the MRI-based only methods and also achieved state-of-the-art performance while using multi-modal data. 
% Our proposed TriFormer achieved the state-of-the-art AUC and accuracy on the ADNI1 and ADNI2 testset, respectively. Compare to the work using only imaging data as single modality input, TriFormer has more than 15\% improvements in predicting \ac{MCI} conversion prediction. We also outperform state-of-the-art works that used the same \ac{MRI} and clinical data.
\begin{table}\label{sota_compare}
\caption{Quantitative comparisons between TriFormer and SOTA methods for predicting MCI to AD conversion on ADNI1 and ADNI2}
\vspace*{1mm}
\resizebox{\columnwidth }{!}{
\begin{tabular}{lSSSSS}
\toprule
  \multirow{2}{*}{Model}&{No. of MCI} &\multicolumn{2}{c}{ADNI1}&\multicolumn{2}{c}{ADNI2}
  % \addlinespace[5pt]
  \\\cmidrule(lr){3-4}\cmidrule(lr){5-6}
  &&{AUC}&{ACC} & {AUC} & {ACC} \\
  \midrule
 {Jang et al. \cite{Jang_2022_CVPR}}& 411& {66.45} & {62.45}  & 73.56 & 72.05 \\
 % {Wen et al. \cite{wen2020convolutional}}& 411& {66.45} & {62.45}  & 73.56 & 72.05 \\
 {Qiu et al. \cite{qiu2020development}} & 411 &{71.82}& {70.11}  & 72.03 & 66.90\\
 {Pan et al. \cite{pan2020spatially}}& 694& {-}& {-}& 83.32 & 77.77\\
 {Lin et al.} \cite{lin2018convolutional}&308  &  86.10 &\textbf{79.9} & {-} & {-}\\
 {Zheng et al. \cite{zheng2022transformer}}& 249 & {-}& {-}  & 88.8 & 83.30\\
 {Guan et al. \cite{guan2021mri}}&455 & {80.80}& {74.60}  & 87.10 & 80.00\\

 \bottomrule
 {TriFormer (proposed)} & 411&\textbf{86.12}  & {77.31} & \textbf{91.47} & \textbf{84.10}\\
\bottomrule
\end{tabular}
}
\end{table}
\subsection{Ablation studies}
We further investigate the effectiveness of the three constituent transformers in TriFormer on the ADNI1 dataset. First, we compare the 2.5D ViT with Wen et al. \cite{wen2020convolutional} which uses a \ac{CNN} as a feature extractor. It improves the ACC of the previous method by 3.45\%, which demonstrates the advantage of using a transformer to extract multi-view features. Then we show that our clinical transformer can better capture the correlation between multi-modal clinical data compared to the \ac{MLP} classifier. Further, we apply both imaging data and clinical data as multi-modal inputs associated with a \ac{MLP} classifier to demonstrate the usage of multi-modal data outperforms single-modality inputs. We finally demonstrate that the modality fusion transformer with the multi-plane information and extracted multi-modal clinical information from the clinical transformer can further boost the performance compared to the normal \ac{MLP} classifier in both accuracy and AUC metrics. The full ablation study is shown in Table 3.

% We performed an ablation study on TriFormer to demonstrate the added value of the different blocks shown in Table \ref{ablation}. We firstly illustrate that adding clinical information to the single \ac{MRI} modality can improve prediction accuracy. Then the usage of a clinical transformer outperforms using MLP to extract multi-modal clinical information. 

% Second version of table, with booktabs.
\begin{table}\label{ablation}
\caption{Ablation study on TriFormer modules}
\vspace*{1mm}
\resizebox{\columnwidth}{!}{
\begin{tabular}{clccc}\toprule
& \multicolumn{2}{l}{Models} & \multicolumn{2}{c}{Metrics}
\\\cmidrule(lr){1-3}\cmidrule(lr){4-5}
Image & Clinical Data & Fusion    & AUC & ACC \\\midrule
3D CNN \cite{wen2020convolutional} & \xmark & \xmark & 65.98 & 70.69 \\
% DenseNet121 & \xmark & \xmark & 76.24 & 79.46 \\
% Jang et al. \cite{Jang_2022_CVPR}& \xmark & \xmark & 73.56 & 72.05\\
2.5D ViT & \xmark & \xmark & 71.88 & 74.14\\
\xmark & MLP & \xmark  & 90.62& 81.61\\
\xmark & Transformer & \xmark   & 91.07& 82.37\\
2.5D ViT & Transformer & MLP   & 91.21& 82.57\\
2.5D ViT & Transformer & Transformer   & \textbf{91.47}& \textbf{84.10}\\\bottomrule
\end{tabular}
}
\end{table}

\section{CONCLUSION}
In this work, we propose a novel transformer-based network, TriFormer, which predicts the conversion of \ac{MCI} using multi-modal data including 3D \ac{MRI} and multi-modal clinical data. The TriFormer effectively extracts multi-view image features and also captures the correlation between different clinical modalities using two of the three transformers. Then, with a third modality fusion transformer with plane positional information, the framework as a whole can better utilize both imaging and clinical features and a normal MLP classifier. Since medical image analysis makes constant use of multi-modal data, this work could benefit other medical research fields that require multi-modal data.

\section{Acknowledgments}
\label{sec:acknowledgments}
This research was supported by the Medical Research Future Fund (MRF1201961) and the Motor Accident Insurance Commission (MAIC), The Queensland Government, Australia (grant number: 2014000857). We would like to thank the Alzheimer’s Disease Neuroimaging Initiative (ADNI) for data collection and sharing.

\bibliographystyle{IEEEbib}
\bibliography{refs}
\acrodef{ADNI}{Alzheimer’s Disease Neuroimaging Initiative}
\acrodef{CDRSB}{Clinical Dementia Rating Sum of Boxes}
\acrodef{ADAS}{Alzheimer's disease assessment scale}
\acrodef{MMSE}{Mini-Mental State Exam}
\acrodef{RAVLT}{Rey Auditory Verbal Learning Test}
\acrodef{AD}{Alzheimer's disease}
\acrodef{CN}{Congnitively normal}
\acrodef{sMCI}{stable MCI}
\acrodef{pMCI}{progressive MCI}
\acrodef{MCI}{Mild cognitive impairment}
\acrodef{MRI}{Magnetic resonance imaging}
\acrodef{PET}{Positron emission tomography}
\acrodef{ViT}{Vision Transformer}
\acrodef{CNN}{Convolutional neural network}
\acrodef{MLP}{multi-layer perceptual}
\end{document}